\documentclass[11pt]{article}
\usepackage[margin=1in]{geometry}
\usepackage{amsmath, amssymb, amsthm}
\usepackage{graphicx}
\usepackage{hyperref}
\usepackage{authblk} 

\usepackage{titlesec}
\usepackage[utf8]{inputenc} 
\usepackage[T1]{fontenc}    
\usepackage{hyperref}      
\usepackage{url}           
\usepackage{booktabs}      
\usepackage{amsfonts}      
\usepackage{nicefrac}      
\usepackage{microtype}      %
\usepackage{lipsum}
\usepackage{graphicx}

\usepackage[vlined,algoruled,linesnumbered]{algorithm2e}
\usepackage{amsmath}
\usepackage{booktabs}
\usepackage{makecell}
\usepackage{float}
\usepackage{array}
\usepackage{rotating}
\usepackage{bm} 
\usepackage{hyperref}
\usepackage[dvipsnames]{xcolor}
\usepackage{ulem}
\usepackage{subcaption}
\usepackage{caption}
\usepackage{titlesec}
\usepackage{bbm}

\newcommand{\Tr}{\operatorname{Tr}}

\renewcommand{\thefootnote}{\fnsymbol{footnote}}

\title{Fill the Gap: Quantifying and Reducing the Modality Gap in Image-Text Representation Learning}

\author[1,*]{François Role}
\author[2]{Sébastien Meyer}
\author[3]{Victor Amblard}

\affil[1]{Université Paris-Cité\\\texttt{francois.role@u-paris.fr}}
\affil[2]{Pôle d'Expertise de la Régulation Numérique (PEReN)\\
}
\affil[3]{Pôle d'Expertise de la Régulation Numérique (PEReN)\\\texttt{victor.amblard@peren.gouv.fr}}

\affil[*]{Corresponding author}

\date{}

\begin{document}
\maketitle

\setcounter{footnote}{0}
\renewcommand{\thefootnote}{\arabic{footnote}}

\begin{abstract}
Vision-language models (VLMs) allow to embed texts and images in a shared representation space. However, it has been shown that these models are subject to a “modality gap phenomenon” meaning there exists a clear separation between the embeddings from one modality and another in the embedding space. While this misalignment is detrimental for downstream tasks  such as multimodal retrieval, multimodal clustering or zero-shot classification, etc. no generic and practical methods have so far been proposed to assess it precisely and even reduce it. We therefore propose novel measures and effective techniques (spectral- and optimal transport-based methods) to achieve this goal. Extensive experiments conducted on several image-text datasets and models demonstrate their effectiveness and beneficial effects on downstream tasks.  Our code is available at~\url{ https://code.peren.gouv.fr/open-source/modeles-multimodaux/article-modeles-multimodaux.git}.

\end{abstract}

\section{Introduction}
Information useful for answering a query or performing a task very often comes in more than one modality (e.g. text and image), which has prompted the need for multi-modal systems capable of searching data irrespective of its nature. Central to these applications is the ability to represent heterogeneous data in a shared latent space, using well-aligned representations that allow to match related items regardless of their source (e.g. matching related texts and images).

Over the last few years, several families of models have emerged to create aligned multimodal embeddings from textual and visual features. In the wake of the seminal CLIP model~\cite{radford_21}, vision-language models (VLMs) have revolutionized the field of multimodal retrieval. A VLM typically combines an image encoder and a text model that are trained so that the image encoder’s representation of a given item is similar to that of a text encoder. More precisely, training is conducted on large sets of image-text pairs, by minimizing either the softmax-based contrastive loss~\cite{radford_21} or the sigmoid loss~\cite{zhai_2023}.

Recently, a new breed of VLMs has emerged with a view to better leveraging the power of large language models (LLMs). One\textbf{} of these models is LLM2CLIP~\cite{huang_2024} where the small text encoder used in the original CLIP architecture is replaced by an LLM pretrained on extensive text corpora. Another recent trend is to use VLMs in  the context of Multimodal Retrieval-augmented generation (MM-RAG). The idea is to rely on popular vision-and-language generative models such as Gemini-Pro, GPT-4V, Llava, MiniCPM-V, Qwen-VL, Gemma 3, etc. to produce contextualized image patch embeddings that can then be matched with corresponding text embeddings obtained by feeding texts (e.g. queries) to the same vision-and-language generative model. This is the approach promoted by models such as Colpali~\cite{faysse_2024} in order to bridge the gap between the text and image modalities.

However, despite being such a key performance factor, the precise quality of the aligned representations is not evaluated directly in most studies, but can only be inferred through the results observed on downstream-tasks. This is one of the motivations for the present work. We argue that both the nature and the quality of the obtained alignment deserve study on their own. In fact, although the vision-language models are supposed to produce semantically well-aligned image and text representations, it appears that the promise is not completely fulfilled, even with recent models such as LLM2CLIP~\cite{huang_2024}, as will be shown in the Experiments section. As an illustrative example, Table \ref{tab:pca} shows the PCA and UMAP visualizations of a large number of image and text embeddings generated by several of the above-mentioned models. As can be seen, these scatter plots clearly highlight a bias related to modality differences: image embeddings and text embeddings are located in different parts of the latent space.

\begin{table}[H]
\centering
\renewcommand{\arraystretch}{0.3}
\begin{tabular}{@{}|c|c|c|@{}}
\hline
& \textbf{PCA} & \textbf{UMAP} \\ \hline

CLIPL14 & \begin{minipage}[c]{0.35\textwidth}
    \centering
    \includegraphics[width=\linewidth]{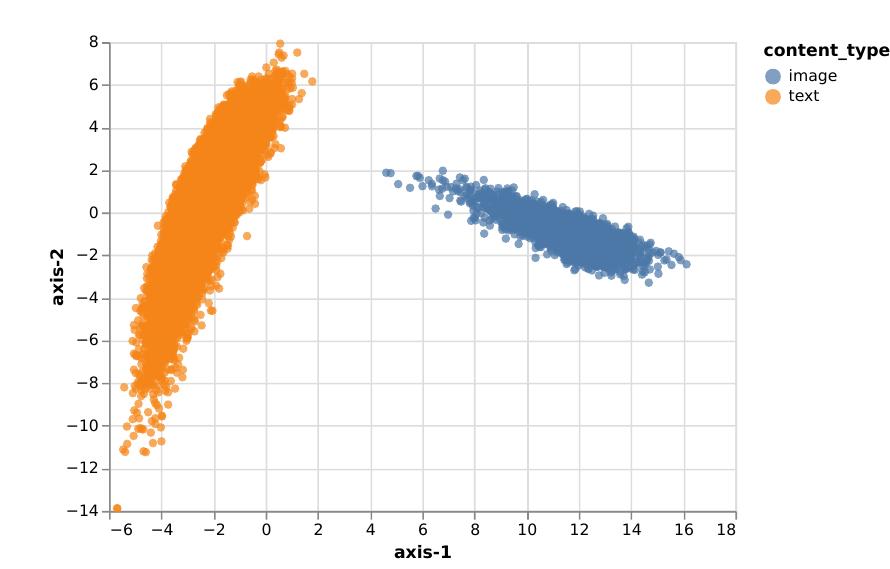}
\end{minipage} & 
\begin{minipage}[c]{0.35\textwidth}
    \centering
    \includegraphics[width=\linewidth]{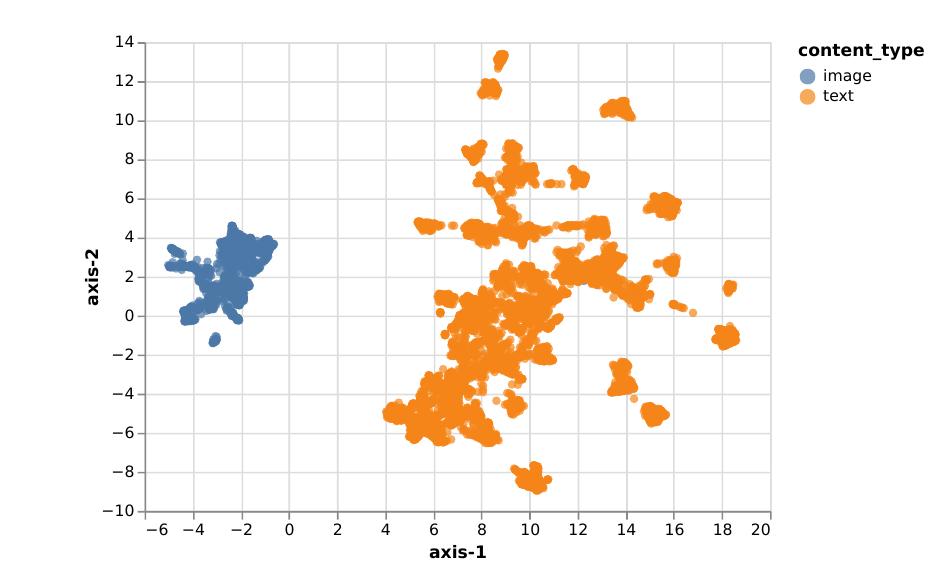}
\end{minipage} \\ \hline

SIGLIP & \begin{minipage}[c]{0.35\textwidth}
    \centering
    \includegraphics[width=\linewidth]{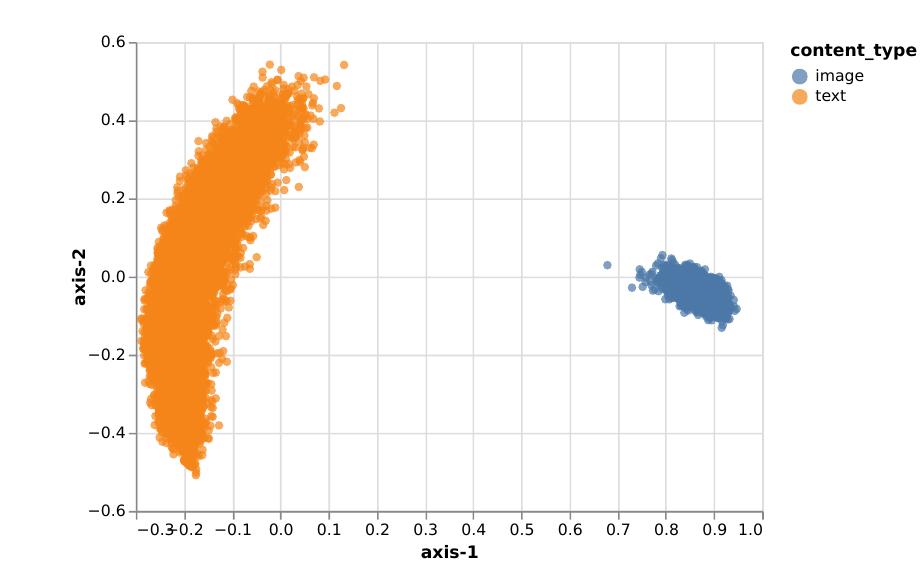}
\end{minipage} & 
\begin{minipage}[c]{0.35\textwidth}
    \centering
    \includegraphics[width=\linewidth]{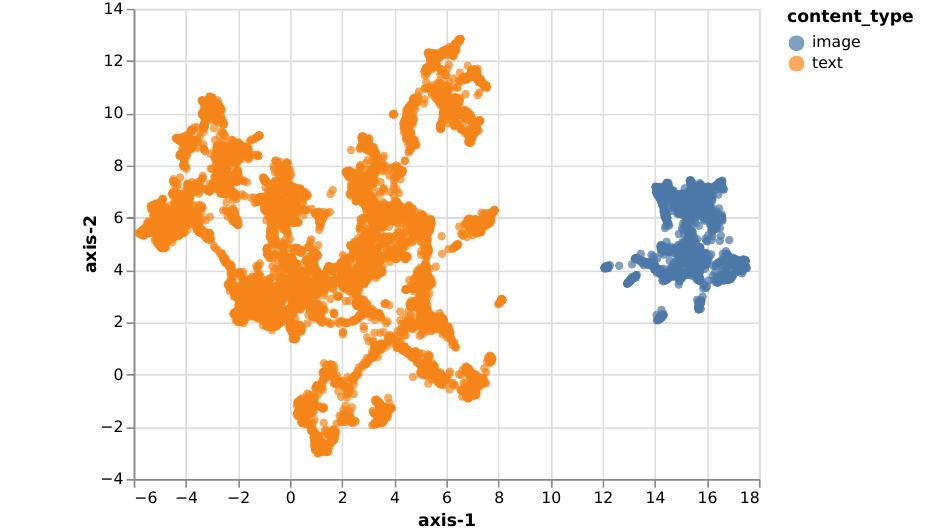}
\end{minipage} \\ \hline

LLM2CLIP & \begin{minipage}[c]{0.35\textwidth}
    \centering
    \includegraphics[width=\linewidth]{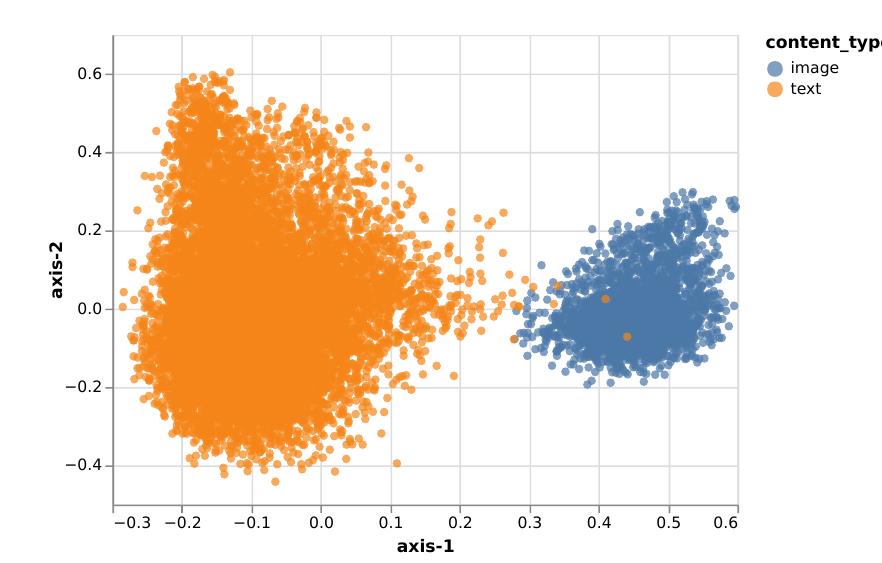}
\end{minipage} & 
\begin{minipage}[c]{0.35\textwidth}
    \centering
    \includegraphics[width=\linewidth]{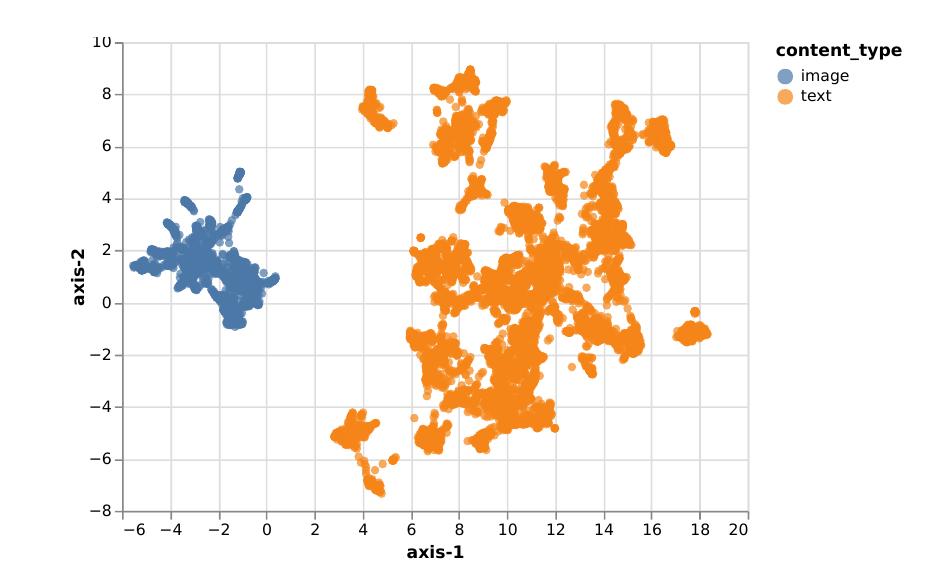}
\end{minipage} \\ \hline

\end{tabular}
\caption{PCA and UMAP visualization of the image (red color) and text (blue) embeddings produced by  different popular models for a random sample of 2500 image-text pairs from the COCO Captions dataset.}
 \label{tab:pca}
\end{table}

This modality gap has already been identified in the context of the CLIP model, and attributed to the combined effects of model initialization and contrastive learning optimization~\cite{liang_2022}. However more research is needed for other kinds of models, and crucially, from a practical perspective, for generic methods allowing to mitigate the gap.  
Indeed, the modality difference bias precludes the embeddings generated by most VLMs from being used in their raw form in fully multimodal information retrieval scenarios, meaning scenarios where the results are expected to include items from various modalities irrespective of the modality of the query. As will be shown in the Experiments section, when querying a dataset containing both images and texts the results
are clearly biased toward the query’s modality. If we query with an image (rep.
a text), the top ranked results are images (resp. texts).

Curiously, while being able to assess and reduce the modality gap is therefore an important issue for vision language models, there is limited literature to report on the topic. The authors in~\cite{liang_2022} propose that it is possible to reduce the gap by fine-tuning a pretrained CLIP model with high temperature, but this claim is specific to CLIP, requires fine-tuning a model and is supported by only a few experiments since, as stated by the authors themselves "The goal of our paper is not to propose a method to close the gap and to improve downstream performance... Systematic analysis of the impact of the gap on applications is an important direction of future work."~\footnote{In fact, the title of our paper echoes a response to the title of the paper in~\cite{liang_2022}}. More recent work~\cite{wei_2024} proposes mitigating the modality bias by designing modified versions of the CLIP and BLIP~\cite{li_2022} models capable of producing unified vectors fusing the text and image information. They do so by augmenting the original architectures of these models with merge or cross-attention layers, which implies quite heavy adaptation for each model.
This motivated us to propose measures and methods generic and applicable to all types of models.

Our contribution is two-fold. First, we present techniques for modifying the out-of-the box representations provided by the multimodal models so that they are better aligned (section~\ref{sec:transformation}).
Second, we propose dedicated metrics for quantifying the alignment quality (section~\ref{sec:measures}). In particular, these measures are used in the experiments of Section ~\ref{sec:experiments} where the transformation methods of Section~\ref{sec:transformation} are shown to be effective.

\section{Methods}\label{sec:methods}

In the following, we assume an image embedding matrix $\bm{X}$ and a text embedding matrix $\bm{Y}$.

\subsection{Post-processing} \label{sec:transformation}

In this section, we propose two families of techniques designed to alleviate the modality gap between text and image. They can be used as a post-processing step to improve the original alignment.

We experimented with two approaches, firstly spectral techniques and secondly optimal transport.

\subsubsection{Spectral Techniques}

Assume an ($n \times n$) weighted adjacency matrix $\mathbf{W}$ of a graph $G$, where the weights represent the similarities between $n$ objects (e.g. $n$ images). We also define $\mathbf{D}$ as the diagonal degree matrix $d_i=\sum_{j=1}^{n}w_{ij}$.
Spectral embedding techniques~\cite{luxburg:2007} are especially well-suited to represent the nodes of $G$  so that two similar nodes $i$ and $j$ (according to the graph weights) will be represented by two vectors that are close in a low-dimensional embedding space.

For example, if we want to project the weighted graph onto a line we have to find the vector $\mathbf{f}$ minimizing $\mathbf{f}^T\mathbf{L}\mathbf{f}=\sum_{i,j=1}^{n}w_{ij}(f_i - f_j)^2$ where $\mathbf{L}=\mathbf{D}-\mathbf{W}$ is the Laplacian matrix, and $\mathbf{f}$ is the vector whose component $f_i$ is the coordinate of the $i$-th vertex\footnote{We have omitted the constraints $\mathbf{f}^T\mathbf{f}=1$ and $\mathbf{f}^T1=0$ from the formula for the sake of clarity.}. This vector $\mathbf{f}$ is the vector corresponding to the smallest non-null eigenvalue of $\mathbf{L}$.

More generally, if we want to embed the nodes in a $k$-dimensional ($1<k$) the solution is the matrix $\mathbf{F}$ whose column vectors $\mathbf{f_i}$ are the eigenvectors corresponding to the $k$ lowest nonzero eigenvalues of $\mathbf{L}$.

To adapt spectral embedding techniques to an image-text scenario, we need to perform the following steps:

\begin{enumerate}
\item Given  the image embedding matrix $\textbf{X}$ and the text embedding matrix $\textbf{Y}$ obtained using a VLM, compute $\textbf{W}=\textbf{XY}^T$,
\item Form the matrix  $\textbf{A}  = \begin{bmatrix} 0 & \textbf{W} \\ \textbf{W}^T & 0 \end{bmatrix}$,
\item Compute the matrix $\textbf{F}= [\textbf{u}_{1}, \dots, \textbf{u}_{k}]$  where the $u_i$ are the eigenvectors corresponding to the $k$ smallest eigenvalues of $\textbf{L}_{rw} = \textbf{D}^{-1} \textbf{L}$, with $\textbf{L}=\textbf{D}-\textbf{A}$ and $\textbf{D}$ being the degree matrix,
\item Use the $i$-th row of $\textbf{F}$ as the representation of the $i$-th point (be it an image or a text).
\end{enumerate}

\subsubsection{Optimal Transport}

Optimal Transport denotes a set of methods aiming to find a transport $T$ that moves a source distribution $\mu$ to a target distribution $\nu$ while optimizing a transport cost $c$ such that $\nu(B) = \mu(T^{-1}(B))$ for any $\nu$-mesurable set $B$. 
Typically the cost function $c$ embeds a regularization function that helps alleviate the computational cost of the optimization or allows to enforce constraints on the transportation plan.

Optimal transport is a natural solution to the alignment problem as we can easily model the two modalities, texts and images, using two discrete probability distributions in the latent space.

Multiple regularization schemes can be used such as the \textit{entropic} or the \textit{Laplacian} regularization. We hereafter focus on the latter, that was initially propsoed by Ferradans et al.\cite{ferradans2014regularized} for color transfer in images and was later taken up in Flamary et al.\cite{flamary2014optimal}. It allows to enforce two constraints on the positions and the displacements of source points so that similar source points have similar displacements and similar positions.

If we consider a simple cost function such as the euclidean distance, the optimization problem can then be written as follows:

\begin{equation}
\begin{split}
\gamma_0 &= \underset{\gamma \in \mathcal{P}}\arg\min\langle \gamma, C\rangle_F + \frac{\eta}{n^2}\left(\lambda_s \sum_{i,j} S_{ij}^s || \gamma \mathbf{x_i} - \gamma \mathbf{x_j} ||^2 \right. \\ 
& \left. + \lambda_t \sum_{i,j} S_{ij}^t || \gamma^T \mathbf{y_i} - \gamma^T \mathbf{y_j} ||^2\right).
\end{split}
\end{equation}
where
\begin{equation}
\begin{split}
\mathcal{P} = \left\{\gamma \in (\mathbb{R}^+)^{n \times n} | \gamma \mathbbm{1}_n = \mu, \gamma^T \mathbbm{1}_N = \nu\right\}.
\end{split}
\end{equation}

where $x_i$ represent the image embeddings (source distribution) and $y_i$ the text embeddings (target distribution) and the transportation plan is written as $T(x)=\gamma x$ and $T^{-1}(y) = \gamma^T y$. $C$ is the cost function and typically here $S_{ij}^s$ (resp. $S_{ij}^t$) denotes the source (resp. target) similarity matrix. Finally, $\lambda_S$ and $\lambda_t$ are two hyperparameters that affect the amount of regularization.

\subsection{Evaluation Measures}\label{sec:measures}

\subsubsection{Distance-based Evaluation Measures}

The most obvious way of measuring how well-aligned the representations is to compute the mean of the norms or squared norms of the differences:

$$
\frac{1}{n} \sum_{i=1}^n \|\bm{x}_i - \bm{y}_i\|^2 = \frac{1}{n} \sum_{i=1}^n \sum_{j=1}^m (x_{ij} - y_{ij})^2 = \frac{1}{n}\|\bm{X} - \bm{Y}\|_F^2 
$$

However, if the value scales of the original matrices and transformed matrices are very different, comparisons based on Euclidean distances may be misleading. In this case, it is more meaningful to use normalized Euclidean distances or scale-invariant metrics such as cosine distance  $1 - \frac{\mathbf{x} \cdot \mathbf{y}}{\|\mathbf{x}\| \|\mathbf{y}\|}$ between two vectors \textbf{x} and \textbf{y}. Once the distances have been brought to comparable scales, the significance of the reduction in difference can be confirmed or refuted by a paired t-test or Wilcoxon test.

\subsubsection{IR-based Heterogeneity Indices}\label{sec:het}

The IR-based heterogeneity indices are applicable to measure the level of heterogeneity of the results in a multimodal information retrieval task. Let $q$ be a query embedding (be it a text or an image) and $D$ a vector database of text and image embeddings. Assume $D$ contains $m$ images and $n$ texts relevant to $q$. Ideally, when querying $D$ using $q$ we hope to find the most relevant texts and images to appear first in the ranked list of retrieved results, irrespective of their modalities. However, as said in the introduction, if the modality gap between image and text has not been sufficiently overcome, an image (resp. text) query embedding will have a tendency to be more similar to the image (resp. text) embeddings than to those of the other modality. As a consequence, a number of the $n$ (resp. $m$) relevant items may be badly ranked because of their modality.

One numerical method we experimented to have a better assessment of this modality-related bias was to estimate the probability that the top-ranked item is of the same modality as the query's.

To do so, we first form the matrix $\bm{Z }$ by stacking the $\bm{X}$ and $\bm{Y}$ matrices, and then compute the similarity matrix $\bm{S}$:

$$
\bm{Z} = \begin{bmatrix} 
\bm{X} \\
\bm{Y}
\end{bmatrix}
\quad \text{and} \quad
\bm{S} = \bm{Z} \bm{Z^T}
$$

Next, denoting the set of indices corresponding to images (resp. texts) as $i\_ids$ ($t\_ids$ resp.) we define the measures $\textbf{ITR}$ (image text ratio) and $\textbf{TIR}$ (text image ratio). $\textbf{ITR}$ (resp. $\textbf{TIR}$) assesses to what extend an image query (resp. text query) is biased towards retrieving image results (resp. text results). The formal definitions are given in Figure~\ref{fig:ITR}). The greater these values are, the stronger the bias is.

\begin{figure}[H]
\centering
$$
\begin{aligned}
\text{i\_n\_ids} &= \left\{ \mathop{\arg\max}_j \bm{S}_{i,j} \mid i \in \text{i\_ids} \right\}, \\
\text{t\_n\_ids} &= \left\{ \mathop{\arg\max}_j \bm{S}_{i,j} \mid i \in \text{t\_ids} \right\}, \\
i\_i\_n\_ids &= i\_n\_ids \cap i\_ids, \\
t\_i\_n\_ids &= t\_n\_ids \cap i\_ids, \\
i\_t\_n\_ids &=  i\_n\_ids \cap t\_ids, \\
t\_t\_n\_ids &= t\_n\_ids \cap t\_ids.
\end{aligned}
$$

$$
\begin{aligned}
\textbf{ITR} &= \frac{\left| i\_i\_n\_ids \right|}{\left| i\_t\_n\_ids \right|}, \\
\textbf{TIR} &= \frac{\left| t\_t\_n\_ids \right|}{\left| t\_i\_n\_ids \right|}.
\end{aligned}
$$

\caption{Definition of the heterogeneity indices: $\textbf{ITR}$ and $\textbf{TIR}$.}
\label{fig:ITR}
\end{figure}

In addition, it may also be informative to know the mean rank of the best ranked text (image) when querying using an image (resp. text). We have thus also defined two additional ranking measures. We define the Text Mean Rank or $\textbf{TMR}$ as the average rank of the highest-ranked text when querying with an image.
 
The Image Mean Rank $\textbf{IMR}$ is defined symmetrically for the case when the query is a text (Figure~\ref{fig:TMR}).

\begin{figure}[h!]
\centering
$$
\begin{aligned}
\mathbf{TMR} &= \frac{1}{n} \sum_{i=1}^n rank \; of \; the \; best\; ranked \; text \\
\mathbf{IMR} &= \frac{1}{n} \sum_{i=1}^n  rank \; of \; the \; best \; ranked \; image
\end{aligned}
$$
\caption{Definition of the ranking-based measures: $\mathbf{TMR}$ and $\mathbf{IMR}$.}
\label{fig:TMR}

\end{figure}

\subsubsection{Distribution-based Evaluation Measures}

Another way to evaluate the distance between embeddings from different modalities is by considering a given modality as a distribution of values. Several metrics exist for evaluating the closeness of such distributions. 

\noindent \textbf{Fréchet Inception Distance \cite{heusel2017fid}.} The Fréchet inception distance or \textbf{FID} is a common metric for assessing the quality of generative models such as GANs. Under the assumption that both modalities follow multivariate normal distributions, we have the following formula for the \textbf{FID}:

$$
\mathbf{FID}^2 = || \mu_{\text{I}} - \mu_{\text{T}} ||_2^2 + \Tr \left({\Sigma_{\text{I}} + \Sigma_{\text{T}} - 2 \sqrt{\Sigma_{\text{I}} \Sigma_{\text{T}}}} \right),
$$

where $\mu_{I,T }$ and $\Sigma_{I,T}$ respectively denote the mean and the standard deviation of the distributions
of image and text embeddings.

\section{Experiments}\label{sec:experiments}

The goal of the presented 
experiments is to assess the effectiveness of the transformation methods described in section~\ref{sec:transformation}.To do so, we compute the measures presented in section~\ref{sec:measures} for a combination of datasets, models and transformation methods.

The dataset management and embedding extraction process have been performed using an \href{https://code.peren.gouv.fr/open-source/benchmark-modeles-multimodaux/}{open source Python package} developed by the authors and other contributors from PEReN (Pôle d'expertise de la régulation numérique).This package implements abstractions that allow to easily add new models or datasets for evaluation evaluation purposes.

In the experiments, we used an image embedding matrix $\bm{X}$ and a text embedding matrix $\bm{Y}$, both of shape $(n \times d)$, where $n$ is the number of items and $d$ is the latent dimension. Each row $\bm{x}_i$ is the image vector representations  of the $i$-th item and row $\bm{y}_i$ its corresponding text vector representation.

\subsection{Datasets and Models Used}

\subsubsection{Datasets}

\noindent \textbf{COCO-PAIRS} is a set of 2500  image-text pairs randomly extracted from the \textit{MSCOCO} dataset~\cite{lin_2014}. \textit{MSCOCO} provides annotations for different classification and image recognition tasks, such as image classification and semantic segmentation. It was first released in 2014 by Microsoft Research. It is made up of more than 80k training images and more than 40k images for both validation and testing.

\vspace{0.25cm}
\noindent \textbf{CC-PAIRS} is a set of 2500  image-text pairs randomly extracted from the \textit{Conceptual Captions} dataset~\cite{sharma_2018}. Conceptual Captions consists of about
3.3M 〈image, description〉 pairs harvested from the Web and therefore provide evaluators with a large number of texts and images with a wide variety of styles.

\subsubsection{Models}

\vspace{0.25cm}
In the experiments described below, we used the image and text embeddings provided by the following models:

\vspace{0.25cm}
\noindent \textbf{CLIP}~\cite{radford_21} CLIP is the groundbreaking model that popularized the the development of joint text-image models trained using contrastive learning. Specifically, we used the \texttt{clip-vit-large-patch14} implementation\footnote{https://huggingface.co/openai/clip-vit-large-patch14}, which relies on a ViT-L/14 Transformer architecture as its image encoder.

\vspace{0.25cm}
\noindent \textbf{SigLIP}~\cite{zhai_2023} Like CLIP, SigLIP is a contrastive-learning-based model, but employs a pairwise sigmoid loss instead of a softmax loss, which dispenses with the costly normalization performed both across images and across texts.

\vspace{0.25cm}
\noindent \textbf{LLM2CLIP}~\cite{huang_2024} leverages the powerful text encoding capabilities of LLMs to improve multimodal representation learning.
More precisely, an LLM (e.g. Llama3-8B) is first fine-tuned using a caption contrastive loss, training the model to distinguish between captions of the same image and those of different images. The fine-tuned LLM is then used (in frozen mode) as the text encoder in a CLIP-style training process, the goal being to fine-tune the image encoder so that its image embeddings are aligned with the output text embeddings from the fine-tuned LLM.

\subsubsection{Parameters for Optimal Transport}
Optimal-transport based experiments relied on the "Python Optimal Transport" library and used Laplacian regularization EMDLaplaceTransport function. The regularization parameters were tuned for each model using grid search. For each model, a transport plan was learnt using a dedicated subset of 5000 pairs from COCO dataset.

\subsection{Experimental Results and Discussion}~\label{sec:results}

Using the measures presented in Section~\ref{sec:measures}, we carried out a set of evaluations to assess the effectiveness of the methods from Section~\ref{sec:transformation} in overcoming the modality gap. More precisely, the results of the experiments presented in Section~\ref{sec:bias-reduction} are meant to help assess the extent to which these techniques reduce the distance between the image and text representations. However, it is one thing to  bring these representations nearer to each other, but it is also necessary to check that the right images are brought closer to the right texts. In Section \ref{sec:relevance} we therefore present the results of measurements intended to verify that the texts (resp. images) brought closer to a given image (resp. text) by the methods of Section~\ref{sec:transformation} are indeed those relevant to this image (resp. text).

\subsubsection{Assessing the Reduction in Bias}\label{sec:bias-reduction}

Results from Tables~\ref{tab:heter} and~\ref{tab:imr-tmr} provide insight into the degree to which the methods from section~\ref{sec:methods} succeed in mitigating the bias of images (resp. texts) in favor of images (resp.texts). The \textbf{ITR} (resp. \textbf{TIR}) values in Table~\ref{tab:heter} indicate how many more or fewer images (resp.texts) than texts (resp. images) are retrieved when querying a mixed dataset of images and texts with an image (resp. text) . As said in Section~\ref{sec:het}, the greater these values are, the stronger the bias is.

\begin{table}[H]
\centering
\scriptsize
\setlength{\tabcolsep}{2pt} 
\renewcommand{\arraystretch}{1.0} 
\begin{tabular}{@{}l|p{0.6cm}p{0.6cm}p{0.6cm}p{0.6cm}p{0.6cm}p{0.6cm}p{0.6cm}p{0.6cm}p{0.6cm}p{0.6cm}p{0.6cm}p{0.6cm}p{0.6cm}p{0.6cm}p{0.6cm}p{0.6cm}p{0.6cm}p{0.6cm}p{0.6cm}p{0.6cm}!{\vrule}@{}}
\toprule
\textbf{}  &
\multicolumn{10}{c|}{COCO-PAIRS} &
\multicolumn{10}{c|}{CC-PAIRS} \\
\cmidrule(lr){2-11} \cmidrule(lr){12-21} 
& \multicolumn{2}{c|}{ORIG} & \multicolumn{2}{c|}{OT} & \multicolumn{2}{c|}{SPEC20} & \multicolumn{2}{c|}{SPEC60} & \multicolumn{2}{c|}{SPEC120} &
  \multicolumn{2}{c|}{ORIG} & \multicolumn{2}{c|}{OT} & \multicolumn{2}{c|}{SPEC20} & \multicolumn{2}{c|}{SPEC60} & \multicolumn{2}{c|}{SPEC120} \\
\cmidrule(lr){2-11} \cmidrule(lr){12-21} 
& \rotatebox{90}{ITR} & \rotatebox{90}{TIR} & \rotatebox{90}{ITR} & \rotatebox{90}{TIR} & \rotatebox{90}{ITR} & \rotatebox{90}{TIR} & \rotatebox{90}{ITR} & \rotatebox{90}{TIR} & \rotatebox{90}{ITR} & \rotatebox{90}{TIR} &
\rotatebox{90}{ITR} & \rotatebox{90}{TIR} & \rotatebox{90}{ITR} & \rotatebox{90}{TIR} &
\rotatebox{90}{ITR} & \rotatebox{90}{TIR} & \rotatebox{90}{ITR} & \rotatebox{90}{TIR} & \rotatebox{90}{ITR} & \rotatebox{90}{TIR} \\ 
\midrule
CLIP & $+\infty$ & 415.66 & 10.73 & 3.53 & 1.88 & 2.31 & 2.52 & 3.74 & 2.28 & 3.53 & 2499 & $+\infty$ & 6.76 & 72.52 &2 & 2.66 & 2.25 & 2 & 1.94 & 1.55 \\
\bottomrule
SigLIP & $+\infty$  & $+\infty$  & 14 & 2.85 & 1.3 & 2 & 1 & 2.5 & 1.1 & 2.98 & $+\infty$ & $+\infty$ & 9.41 &118 & 1.95 & 2 & 1.78 & 1.16 & 1.58 & 0.84  \\
\bottomrule
LLM2CLIP & 832.33 & 207.33 & 12.15 & 2.93 & 1.17 & 1.48 & 0.82 & 1.58 & 0.85 & 1.55 & 21.93 & 29.86 & 21.72 & 38.68 & 1 & 1.59 & 0.57 & 0.74 & 0.46 & 0.56 \\
\bottomrule
\end{tabular}
\caption{Heterogeneity indices. ORIG and OT denote the original embeddings and the embeddings post-processed using Optimal Transport resp. SPEC20, SPEC60 and SPEC120 stand for the spectral representations with $20$, $60$ and $120$ components resp. }
\label{tab:heter}
\end{table}

\begin{table}[H]
\centering
\scriptsize
\setlength{\tabcolsep}{2pt}
\renewcommand{\arraystretch}{1.0} 
\begin{tabular}{@{}l|p{0.6cm}p{0.6cm}p{0.6cm}p{0.6cm}p{0.6cm}p{0.6cm}p{0.6cm}p{0.6cm}p{0.6cm}p{0.6cm}p{0.6cm}p{0.6cm}p{0.6cm}p{0.6cm}p{0.6cm}p{0.6cm}p{0.6cm}p{0.6cm}p{0.6cm}p{0.6cm}|@{}}
\toprule
\textbf{}  &
\multicolumn{10}{c|}{COCO-PAIRS} &
\multicolumn{10}{c|}{CC-PAIRS} \\
\cmidrule(lr){2-11} \cmidrule(lr){12-21} 
& \multicolumn{2}{c|}{ORIG} & \multicolumn{2}{c|}{OT} & \multicolumn{2}{c|}{SPEC20} & \multicolumn{2}{c|}{SPEC60} & \multicolumn{2}{c|}{SPEC120} &
  \multicolumn{2}{c|}{ORIG} & \multicolumn{2}{c|}{OT} & \multicolumn{2}{c|}{SPEC20} & \multicolumn{2}{c|}{SPEC60} & \multicolumn{2}{c|}{SPEC120} \\
\cmidrule(lr){2-11} \cmidrule(lr){12-21} 
& \rotatebox{90}{IMR} & \rotatebox{90}{TMR} & \rotatebox{90}{IMR} & \rotatebox{90}{TMR} & \rotatebox{90}{IMR} & \rotatebox{90}{TMR} & \rotatebox{90}{IMR} & \rotatebox{90}{TMR} & \rotatebox{90}{IMR} & \rotatebox{90}{TMR} &
\rotatebox{90}{IMR} & \rotatebox{90}{TMR} & \rotatebox{90}{IMR} & \rotatebox{90}{TMR} &
\rotatebox{90}{IMR} & \rotatebox{90}{TMR} & \rotatebox{90}{IMR} & \rotatebox{90}{TMR} & \rotatebox{90}{IMR} & \rotatebox{90}{TMR} \\ 
\midrule
CLIP & 732 & 2474 & 3 & 8 & 4 & 3 & 5 & 4 & 5 & 4 & 1820 & 1976 & 106 & 8 & 4 & 3 & 3 & 4 & 3 & 4 \\
\bottomrule
SigLIP & 1061 & 2500 & 3 & 11 & 4 & 3 & 4 & 2 & 5 & 2 & 1028 & 2498 & 93 & 37 & 3 & 4 & 2 & 4 & 2 & 4  \\
\bottomrule
LLM2CLIP & 133 & 96 & 3 & 8 & 3 & 2 & 3 & 2 & 3 & 2 & 31 & 32 & 40 & 20 & 3 & 2 & 2 & 2 & 2 & 2 \\
\bottomrule
\end{tabular}
\caption{IMR and TMR (values rounded to the nearest integer).}
\label{tab:imr-tmr}
\end{table}

The \textbf{IMR} (resp.\textbf{TMR}) values in Table~\ref{tab:imr-tmr} give the average rank of the highest-ranked texts (resp. images) when querying a mixed dataset of images and texts with an image (resp. text).

\begin{table}[H]
\centering
\scriptsize
\setlength{\tabcolsep}{2pt} 
\renewcommand{\arraystretch}{1.0} 
\begin{tabular}{@{}l|>{\centering}p{1.2cm}>{\centering}p{1.2cm}>{\centering}p{1.2cm}>{\centering}p{1.2cm}>{\centering}p{1.2cm}>{\centering}p{1.2cm}>{\centering}p{1.2cm}>{\centering}p{1.2cm}>{\centering}p{1.2cm}p{1.2cm}|@{}}
\toprule
\textbf{}  &
\multicolumn{5}{c|}{COCO-PAIRS} &
\multicolumn{5}{c|}{CC-PAIRS} \\
\cmidrule(lr){2-6} \cmidrule(lr){7-11} 
& \multicolumn{1}{c|}{ORIG} & \multicolumn{1}{c|}{OT} & \multicolumn{1}{c|}{SPEC20} & \multicolumn{1}{c|}{SPEC60} & \multicolumn{1}{c|}{SPEC120} &
  \multicolumn{1}{c|}{ORIG} & \multicolumn{1}{c|}{OT} & \multicolumn{1}{c|}{SPEC20} & \multicolumn{1}{c|}{SPEC60} & \multicolumn{1}{c|}{SPEC120} \\
\cmidrule(lr){2-6} \cmidrule(lr){7-11} 

CLIP & 287.1 & 35.41 & $\approx 0$ & $\approx 0$ & $\approx 0$ & 280.86 & 92.15 & $\approx 0$ & $\approx 0$ & $\approx 0$  \\
\bottomrule
SigLIP & 1.39 & 0.15 & $\approx 0$ &$\approx 0$ & $\approx 0$& 1.35 & 0.36 &$\approx 0$ &$\approx 0$ & $\approx 0$   \\
\bottomrule
LLM2CLIP & 0.72 & 0.34 &$\approx 0$ & $\approx 0$ & $\approx 0$ & 0.57 & 0.78 & $\approx 0$ & $\approx 0$ & $\approx 0$  \\
\bottomrule
\end{tabular}
\caption{FID.}
\label{tab:fid}
\end{table}

The obtained results clearly suggest that the methods from Section~\ref{sec:transformation} succeed to a large extent in overcoming the modality bias. As will be confirmed in Section~\ref{sec:relevance} the drop in \textbf{IMR} and \textbf{TMR} values is due to the fact that in their transformed forms (OT, SPEC20, SPEC60, SPEC120) the image (resp. text) embeddings have been brought closer to their corresponding text (resp. image) embeddings. With these representations, an image (resp. a text) may be among the top-ranked items when the mixed dataset of text and images is queried with a text (resp. an image). This is in sharp contrast to what we observe when using the original embeddings produced by the models. In the case of CLIP and SigLIP, many \textbf{IMR} and \textbf{TMR} values are close to $1000$ with peaks up to 2500. Recalling that each of our datasets consists of $2500$ image-text pairs, this means that when querying with an image (resp. text), using the original embeddings, almost all images (resp. texts) are ranked before the texts (resp. images). Incidentally, it is worth noting that the \textbf{IMR} and \textbf{TMR} (using the original embeddings) values are markedly lower for the LLM2CLIP model, which is in accordance with what can be seen at the bottom of Table~\ref{tab:pca} and reflects the influence of contextualization. Nevertheless, the decrease in \textbf{IMR} and \textbf{TMR} values is also very significant in this case.

Finally, the distances are also significantly reduced as shown by Table~\ref{tab:fid} and Figure~\ref{fig:distances}.

\subsubsection{Assessing the Relevance of the Alignment Procedure}~\label{sec:relevance}

 As said in the introduction of Section~\ref{sec:results}, it is not enough to reduce the average distance between image and text representations uniformly. One must also assess the ability of the methods from Section~\ref{sec:methods} to bring an image (resp. text) closer to its related text (resp. image)  compared to the image's (resp. text's)  proximity to all other texts (resp. images).
We compute the distances between the images and their corresponding texts in both cases (original  and transformed embeddings) using the cosine distance $1 - \frac{\mathbf{x} \cdot \mathbf{y}}{\|\mathbf{x}\| \|\mathbf{y}\|}$ between two vectors \textbf{x} and \textbf{y}.

\begin{figure}[H]
    \centering

    \begin{subfigure}[t]{\textwidth}
        \centering
        \includegraphics[width=0.83\textwidth]{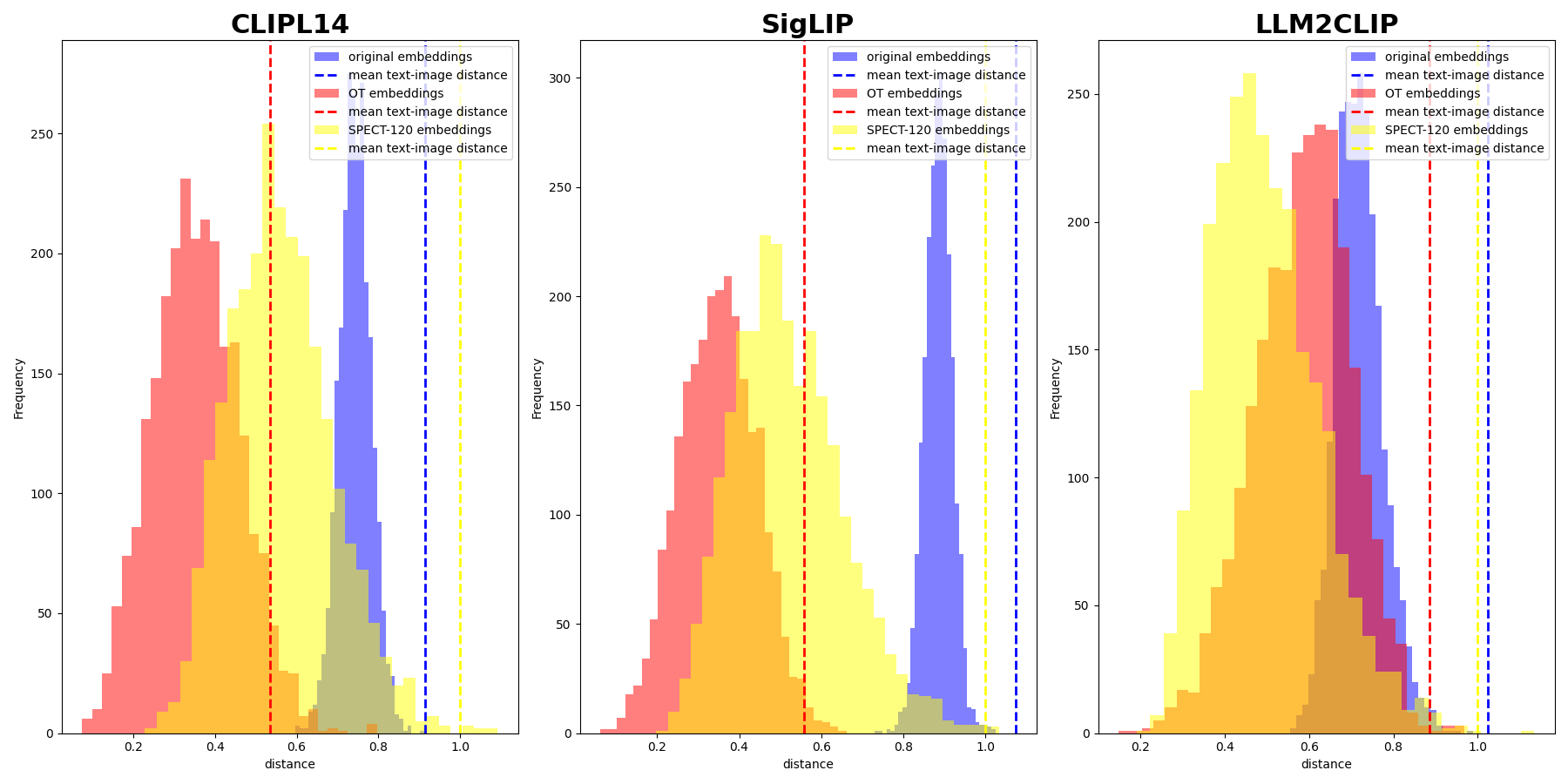}
        
        \caption{COCO-PAIRS}
    \end{subfigure}

    \vspace{1em} 

    \begin{subfigure}[t]{\textwidth}
        \centering
        \includegraphics[width=0.83\textwidth]{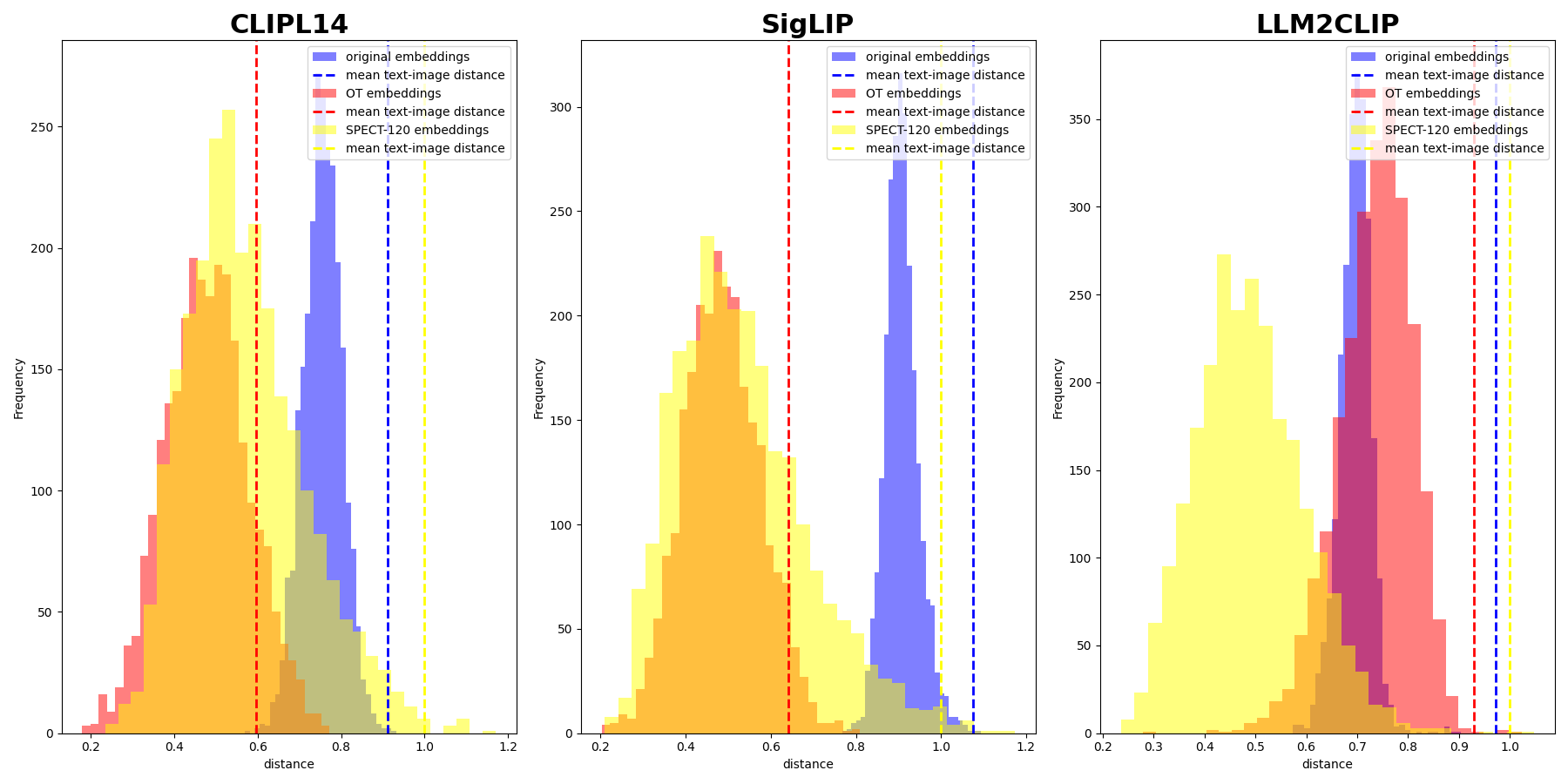}
        
        \caption{CC-PAIRS}
    \end{subfigure}

    \caption{Distribution of the cosine distances between the images and their corresponding texts in both cases (original  and transformed embeddings). The dashed lines represent the mean distances across all image-text pairs.}
    \label{fig:distances}
\end{figure}

We then compare the mean of these distances with the mean distance across all image-text pairs. As can be seen in Figure~\ref{fig:distances}, the former distance is significantly lower than the latter one in all our scenarios (ORIG, OT, SPECT20, SPECT60, SPECT120), but the difference is especially noticeable for the transformed embeddings and particularly for the SPECT embeddings. This shows that the techniques from Section~\ref{sec:methods} not only bring images and texts closer but does so by moving the images (resp. texts) closer to the texts (resp. images) that correspond to them.
This is confirmed by Table~\ref{tab:recall} where we evaluate recall at various levels when querying the datasets with  images, the gold truth being the texts corresponding to these images. It can be seen that when using the original raw embeddings there is almost no chance of retrieving the relevant texts even when considering the top 20 results. In contrast, the transformation methods, especially the spectral ones, greatly enhance recall performance. 
It is also worth noting, that in the case of the COCO dataset we only use a single caption (out of the five possible captions per image provided in this dataset) compared to other more lenient benchmarks where a retrieved result is considered to be correct if it overlaps with any of the relevant (ground-truth) five instances. Since we are using only one ground-truth caption per image, this naturally results in lower but more realistic scores.

\begin{table}[h]
\centering
\scriptsize
\setlength{\tabcolsep}{2pt} 
\renewcommand{\arraystretch}{1.0} 
\begin{tabular}{@{}l|>{\centering}p{0.5cm}>{\centering}p{0.5cm}>{\centering}p{0.5cm}>{\centering}p{0.5cm}>{\centering}p{0.5cm}>{\centering}p{0.5cm}>{\centering}p{0.5cm}>{\centering}p{0.5cm}p{0.5cm}p{0.5cm}p{0.5cm}>{\centering}p{0.5cm}p{0.5cm}>{\centering}p{0.5cm}p{0.5cm}>{\centering}p{0.5cm}>{\centering}p{0.5cm}>{\centering}p{0.5cm}>{\centering}p{0.5cm}p{0.5cm}p{0.5cm}>{\centering}p{0.5cm}>{\centering}p{0.5cm}p{0.5cm}|@{}}
\toprule
\textbf{}  &
\multicolumn{12}{c|}{COCO-PAIRS} &
\multicolumn{12}{c|}{CC-PAIRS} \\
\cmidrule(lr){2-13} \cmidrule(lr){14-25}  
 & \multicolumn{3}{c|}{ORIG} & \multicolumn{3}{c|}{OT} & \multicolumn{3}{c|}{SPEC60} & \multicolumn{3}{c|}{SPEC120} &
  \multicolumn{3}{c|}{ORIG} & \multicolumn{3}{c|}{OT} & \multicolumn{3}{c|}{SPEC60} & \multicolumn{3}{c|}{SPEC120} \\
\cmidrule(lr){2-13} \cmidrule(lr){14-25} 
K & \rotatebox{0}{5} & \rotatebox{0}{10} & \rotatebox{0}{20} & \rotatebox{0}{5} & \rotatebox{0}{10} & \rotatebox{0}{20} & \rotatebox{0}{5} & \rotatebox{0}{10} & \rotatebox{0}{20} & \rotatebox{0}{5} &
\rotatebox{0}{10} & \rotatebox{0}{20} & \rotatebox{0}{5} & \rotatebox{0}{10} &
\rotatebox{0}{20} & \rotatebox{0}{5} & \rotatebox{0}{10} & \rotatebox{0}{20} & \rotatebox{0}{5} & \rotatebox{0}{10} & \rotatebox{0}{20} & \rotatebox{0}{5} & \rotatebox{0}{10} & \rotatebox{0}{20} \\ 
\midrule
CLIP & 0 & 0 & 0 & 0.09 & 0.15 & 0.26 & 0.38 & 0.52 & 0.68 & \textbf{0.45} & \textbf{0.6} & \textbf{0.74} & 0 & 0 & 0 & 0.1 & 0.18 & 0.25 & 0.44 & 0.56 & 0.68 & \textbf{0.56} & \textbf{0.68} & \textbf{0.78} \\
\bottomrule
SigLIP & 0 & 0 & 0 & 0.04 & 0.09 & 0.18 & 0.55 & 0.68 & 0.79 & \textbf{0.58} & \textbf{0.7} & \textbf{0.81} & 0 & 0 & 0 & 0.14 & 0.2 & 0.3 & 0.53 & 0.64 & 0.74 & \textbf{0.6} & \textbf{0.7} & \textbf{0.8}  \\
\bottomrule
LLM2CLIP & 0.01 & 0.03 & 0.08 & 0.17 & 0.29 & 0.46 & 0.63 & 0.74 & 0.85 & \textbf{0.67} & \textbf{0.79} & \textbf{0.87} & 0.2 & 0.38 & 0.58 & 0.16 & 0.3 & 0.48 & 0.81 & 0.89 & 0.94 & \textbf{0.91} & \textbf{0.96} & \textbf{0.98}\\

\bottomrule
\end{tabular}
\caption{Recall at $K$ when querying the mixed datasets (each containing 2500 images and 2500 corresponding texts each) with an image and trying to retrieve the corresponding text caption or description. The SPEC20 case, which shows similar results as the other SPEC cases,  has been left out to keep the table from being too cluttered.}
\label{tab:recall}
\end{table}

Finally, to confirm that correcting the misalignment requires criteria accounting for the similarity between images and texts and not just simple dimension reduction techniques~\footnote{As plots in Table~\ref{tab:pca} had already strongly suggested.}, Table~\ref{tab:recall-pca} show the recall results obtained when performing PCA. As can be seen, the obtained results do not improve over those obtained with the original embeddings,and are even slightly worse.

\begin{table}[H]
    \centering
    \scriptsize
    \begin{tabular}{l|ccc|ccc|ccc|ccc|ccc|ccc|}
    \toprule
    & \multicolumn{9}{c|}{COCO-PAIRS} & \multicolumn{9}{c|}{CC-PAIRS} \\
    \cmidrule(lr){2-10} \cmidrule(lr){11-19}
    & \multicolumn{3}{c|}{PCA20} & \multicolumn{3}{c|}{PCA60} & \multicolumn{3}{c|}{PCA120} & 
      \multicolumn{3}{c|}{PCA20} & \multicolumn{3}{c|}{PCA60} & \multicolumn{3}{c|}{PCA120} \\
    \cmidrule(lr){2-4} \cmidrule(lr){5-7} \cmidrule(lr){8-10} 
    \cmidrule(lr){11-13} \cmidrule(lr){14-16} \cmidrule(lr){17-19}
    K & \rotatebox{0}{5} & \rotatebox{0}{10} & \rotatebox{0}{20} & 
        \rotatebox{0}{5} & \rotatebox{0}{10} & \rotatebox{0}{20} & 
        \rotatebox{0}{5} & \rotatebox{0}{10} & \rotatebox{0}{20} & 
        \rotatebox{0}{5} & \rotatebox{0}{10} & \rotatebox{0}{20} & 
        \rotatebox{0}{5} & \rotatebox{0}{10} & \rotatebox{0}{20} & 
        \rotatebox{0}{5} & \rotatebox{0}{10} & \rotatebox{0}{20} \\
    \midrule
    CLIP & 0 & 0 & 0 & 0 & 0 & 0 & 0 & 0 & 0 & 
          0 & 0 & 0 & 0 & 0 & 0 & 0 & 0 & 0 \\
    \bottomrule
    SigCLIP & 0 & 0 & 0 & 0 & 0 & 0 & 0 & 0 & 0 & 
          0 & 0 & 0 & 0 & 0 & 0 & 0 & 0 & 0 \\
    \bottomrule
    LLM2CLIP & 0 & 0 & 0 & 0 & 0 & 0 & 0.01 & 0.02 & 0.06 & 
          0.01 & 0.02 & 0.03 & 0.05 & 0.1 & 0.17 & 0.14 & 0.25 & 0.38 \\
    \bottomrule
    \end{tabular}
    \caption{Recall results for PCA with 20, 60, and 120 principal components}
    \label{tab:recall-pca}
\end{table}

The main limitations of the proposed methods are the high time complexity of computing the eigenvectors of the Laplacian matrix (for the spectral-based method), and the need to train a model (in the case of the optimal transport-based method). As concerns the spectral method, it should however be noted that since one doesn't need to compute the entire spectrum (good results are already obtained with 60 components and very good ones  with 100), truncated eigenvalue decomposition can be used to reduce computational cost.

\section{Conclusion}

In this paper novel measures as well as spectral- and optimal-transport-based methods have been presented to address the "modality gap" phenomenon encountered in most vision-language models. The aim of these methods is to bring the representations of images (resp. texts) closer to those of texts (resp. images) that correspond to them. We have also proposed new, specialized measures to help precisely assess the level of the gap. By using these measures as part of extensive experiments, we were able to show the effectiveness of the proposed methods, especially the spectral-based ones. In particular, the experiments revealed that a limited number of components (around 100) was sufficient to obtain very good recall results on retrieval tasks involving a mixed dataset of images and texts, results that are completely out of reach if one uses the original embeddings produced by the VLMs even though they have a much larger size. 
Directions for extending this preliminary investigation include integrating additional models and datasets into the experiments, making statistical measures more robust, and using more advanced algorithms for the computation of spectral embeddings.

\section{Acknowledgements}
The authors would like to thank Lucas Serrano and Camilla Penzo from the Pôle d'Expertise de la Régulation Numérique (PEReN) for their contribution to research and development in multimodal systems carried out at PEReN, as well as to the \href{https://code.peren.gouv.fr/open-source/benchmark-modeles-multimodaux/}{benchmark-modeles-multimodaux package}.

\end{document}